\documentclass[runningheads]{llncs}
\usepackage[T1]{fontenc}
\usepackage{newtxtext}       %
\usepackage[varvw]{newtxmath}       
\usepackage{graphicx}
\usepackage{amsmath}
\usepackage{booktabs}
\usepackage{array}
\usepackage{multirow}
\usepackage{xcolor}
\usepackage{algorithm}
\usepackage{algpseudocode}
\usepackage{tikz}
\usetikzlibrary{arrows.meta,positioning,shapes.geometric,calc,fit,backgrounds}
\usepackage{pgfplots}
\pgfplotsset{compat=1.16}
\usepgfplotslibrary{groupplots}
\usepackage{hyperref}
\hypersetup{colorlinks=true, linkcolor=blue, citecolor=blue, urlcolor=blue}

\definecolor{tgblue}{HTML}{2F6DB5}
\definecolor{tggreen}{HTML}{4C9A2A}
\definecolor{tgorange}{HTML}{E8772E}
\definecolor{tgteal}{HTML}{3BB0C9}
\definecolor{tggold}{HTML}{D4A23C}
\definecolor{tgred}{HTML}{C0392B}
\definecolor{tggrey}{HTML}{6B7280}
\newcommand{\E}{\varepsilon}
\algrenewcommand\algorithmicrequire{\textbf{Input:}}
\algrenewcommand\algorithmicensure{\textbf{Output:}}

\newcommand{\tablesetup}{\small\setlength{\tabcolsep}{5pt}\renewcommand{\arraystretch}{1.1}}
\begin{document}
\title{Trade Promotion Optimisation for Consumer Goods}
\titlerunning{Analogue-Pooled Elasticity Curves for Trade Promotion Optimisation}
\author{Ankur Verma\inst{1} \and
Namrata Gupta\inst{2}}
\authorrunning{A. Verma et al.}
\institute{TrueGradient AI, \email{ankur@truegradient.ai} \and
TrueGradient AI, \email{namrata@truegradient.ai}}
\maketitle              
\begin{abstract}
Trade promotions consume a large share of consumer-goods revenue, yet most planners still
set discounts with a single, often unreliable, price elasticity per item. We present an
end-to-end Trade Promotion Optimisation (TPO) system built as one chain: analogue SKUs, point
elasticity, elasticity curve, price recommendation, promotional lift and net lift. Elasticity
is measured at post-promo time from directional price-dip events against a rolling regular
price, and at promo time by a causal estimator that residualises trend, seasonality and
promotion, handles stock-out censoring with a Tobit likelihood and updates itself from the
first clean promotion week. Items with weak evidence borrow the point elasticity and the
curvature from analogue SKUs chosen by a cross-validated ridge model over category, brand,
sub-category, price line and description. The optimiser separates self-price lift from
substitution and post-promo dips and adjusts for inventory and trade funding. In a four-week
promotion experiment on a 39-SKU consumer brand (mean elasticity $-3.06$), the plan lifts units
by 20.6\% and revenue by 6.9\% at flat margin. Analogue pooling
halves elasticity error relative to hierarchy averaging.
\end{abstract}
\section{Introduction}
Trade promotions, temporary price reductions funded jointly by manufacturers and retailers,
are a major consumer-goods spend whose effectiveness is hard to
measure~\cite{blattberg1990,cohen2017}. Choosing between a 20\% and a 30\% discount needs how units respond to price (the
\emph{self-price elasticity}), how much of the extra volume is taken from sibling items
(\emph{substitution}), how much is merely pulled forward from future weeks (the
\emph{post-promo dip}), and whether the stock to serve the uplift exists. All are noisy for a single SKU.

Elasticity is the most influential number in a promotion recommendation, yet the least well
identified: most SKUs have few price events, seasonality confounds naive regressions, and
stock-outs censor demand when discounts are deepest. A linear response then drives elastic
items to the deepest permitted discount, a ``corner solution'' planners distrust.

\begin{figure}[t]
\centering
\begin{tikzpicture}[
  node distance=2.2mm,
  st/.style={draw, rounded corners=2pt, align=center, font=\scriptsize, minimum height=10mm, text width=15mm, fill=#1!12, draw=#1!70!black},
  arr/.style={-{Stealth[length=1.6mm]}, thick, tggrey}]
\node[st=tgorange] (a) {Analogue\\ SKUs};
\node[st=tgblue, right=of a] (b) {Point\\ elasticity $b_1$};
\node[st=tgblue, right=of b] (c) {Curve\\ $e(d){=}b_1{+}b_2d$};
\node[st=tggreen, right=of c] (d) {Price\\ recommend.};
\node[st=tggreen, right=of d] (e) {Promo\\ lift};
\node[st=tgteal, right=of e] (f) {Net\\ lift};
\draw[arr] (a)--(b); \draw[arr] (b)--(c); \draw[arr] (c)--(d); \draw[arr] (d)--(e); \draw[arr] (e)--(f);
\draw[arr, dashed] (a.south) to[out=-25,in=-155] node[below, font=\tiny]{curvature $b_2$} (c.south);
\node[font=\tiny, below=0.5mm of f, text width=17mm, align=center] {$-$ substitution\\ $-$ post-promo dip};
\end{tikzpicture}
\caption{The chain that structures the paper: evidence borrowed from analogue SKUs fixes the
point elasticity and the curvature, the curve sets the recommended discount, and the
recommended discount yields a promotional lift that is netted into incremental value.}
\label{fig:chain}
\end{figure}
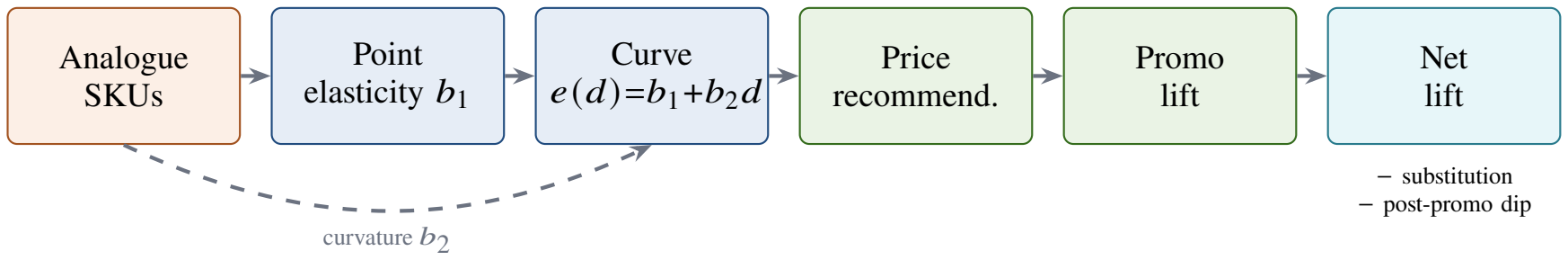

Our central idea is that these problems must be solved as one \emph{chain} rather than as
separate modules (Fig.~\ref{fig:chain}): analogue SKUs supply evidence where an item has
little of its own; that evidence fixes the \emph{point elasticity} $b_1$ and the
\emph{curvature} $b_2$; together they form an \emph{elasticity curve} that bends as the
discount deepens; the curve gives the optimiser an interior \emph{price recommendation};
the recommendation produces a \emph{promotional lift}; and netting out substitution and the
post-promo dip turns it into the \emph{net lift} the business actually earns. We make four
contributions:
\begin{enumerate}
\item A \textbf{two-regime elasticity estimator}: an event-based post-promo-time estimator
measured against a rolling regular price (Sect.~\ref{sec:post}), and a promo-time causal curve
estimator with stock-out-aware Tobit fitting and a Bayesian update from realised promotion
weeks (Sect.~\ref{sec:promotime}).
\item A \textbf{learned analogue-pooling model} that selects donor SKUs by category, brand,
sub-category, price line and description and shrinks each item's own estimate towards a
cross-validated prediction, for both $b_1$ and $b_2$ (Sect.~\ref{sec:analogue}).
\item A \textbf{self- versus cross-price decomposition} that turns gross promotional lift into
net incremental lift after substitution and post-promo dips (Sect.~\ref{sec:cross}).
\item An \textbf{optimiser aware of inventory and funding}, traced end to end on a worked
example (Sect.~\ref{sec:chain}) and evaluated in a four-week promotion experiment on a 39-SKU
brand and in a controlled simulation (Sect.~\ref{sec:exp}).
\end{enumerate}

\section{Related Work}
\label{sec:related}
Cohen et al.~\cite{cohen2017} cast promotion planning as a mixed-integer programme with
past-promotion memory; Ferreira et al.~\cite{ferreira2016} couple demand forecasts with price
optimisation, and Smith and Achabal~\cite{smith1998} price under inventory-dependent demand;
foundations are in~\cite{talluri2004}. Our baseline demand comes from neural forecasters
combined by stacked generalisation over price, promotion, event, holiday and weather
drivers~\cite{tgdemand2020}, and the ML elasticity equations and their optimisers follow~\cite{tgelastic2020}. Promotional bumps have long been decomposed into
switching, acceleration and stockpiling~\cite{gupta1988}, with post-promotion
dips~\cite{vanheerde2000}, only a third of the bump due to brand switching~\cite{vanheerde2003}
and extra consumption~\cite{ailawadi1998}; Bijmolt et al.~\cite{bijmolt2005} report a mean
elasticity of about $-2.6$, and K\"ok and Fisher~\cite{kok2007} model substitution.
Residualising treatment and outcome follows Frisch-Waugh-Lovell~\cite{frisch1933,lovell1963}
and double machine learning~\cite{chernozhukov2018}; censored demand calls for
Tobit~\cite{tobin1958}. Borrowing strength across items draws on empirical
Bayes~\cite{efron1973}, hierarchical models~\cite{gelman2007,montgomery1997} and
comparables~\cite{baardman2018}; our analogue model adds ridge~\cite{hoerl1970} and
TF-IDF~\cite{salton1988} features.

\section{Problem Formulation}
\label{sec:problem}
Let $\mathcal{I}$ be the set of series (SKU, optionally $\times$ store or channel), observed
at weekly periods $t$. For item $i$ we observe units $q_{it}$, selling price $p_{it}$, a
promotion flag $\pi_{it}$, unit cost $k_i$, list price (MRP) $\bar p_i$, stock signals and a
baseline forecast $\hat q_{it}$ for the horizon $\mathcal{H}$, produced by stacked neural demand
models that take price, promotions, events, holidays and weather as inputs~\cite{tgdemand2020}. A promotion sets a relative
price change $c_i\in\mathcal{C}_i\subset[c^{\min},c^{\max}]$ around the regular price
$p^{0}_i$, with discount depth $d_i=-\min(c_i,0)$. We write $\E_i$ for the self-price
elasticity, $e_i(d)=b_{1i}+b_{2i}d$ for the promo-time elasticity curve, $\E_{ji}$ for the
cross-price (substitution) elasticity of $j$ to the price of $i$, and $\E^{\mathrm{post}}_i$
for the post-promo elasticity. The corridor defaults to $[-20,+20]\%$ with step $s=4$ points.
The per-item objective trades revenue gain against margin gain,
\begin{equation}
\max_{c\in\mathcal C_i}\ J_i(c)=\tfrac{1}{100}\Big[w\,\big(R_i(c)-R_i(0)\big)+(100-w)\big(M_i(c)-M_i(0)\big)\Big],
\label{eq:obj}
\end{equation}
where $R_i(c)=p^0_i(1+c)\,Q_i(c)$, $M_i(c)=(p^0_i(1+c)-k_i)\,Q_i(c)$ and the net demand
response $Q_i(c)$ includes self-price lift, substitution and post-promo effects
(Sect.~\ref{sec:cross}). Unit cost is held fixed, so deeper cuts narrow unit margin.

\section{Methodology}
\label{sec:method}
Algorithm~\ref{alg:tpo} implements the chain of Fig.~\ref{fig:chain}: items are processed
in parallel, and only the analogue model and hierarchy statistics are fitted across items.
Each subsection below is one link; Sect.~\ref{sec:chain} follows a single SKU through all of them.

\subsection{Promotion Flag and Regular Price}
\label{sec:base}
A period is promotional if a promo column says so, else if $p_{it}\le0.9\,p^{\mathrm{base}}_{it}$
(client base price, else a units-weighted average).
The \emph{regular price} $p^0_i$ is a client regular-price column when supplied, else the
\emph{modal-above-median} price of the trailing 365 days (most frequent of the five commonest
prices at or above the median, ties to the higher price), so an item on deal today is not
optimised from its promo price.

\subsection{Post-Promo-Time Elasticity from Price Events}
\label{sec:post}
Once promotions have closed we measure elasticity from the price events they created. Per
item, history within a 730-day look-back is aggregated to one row per period and
traversed sequentially while tracking a \emph{rolling regular price}, the mode-above-median
$\tilde p_t$ of a window $\mathcal W_t$ holding the last 12 \emph{regular} prices. A period
opens or continues a \emph{dip} if $p_t\le(1-\theta)\tilde p_t$ and a \emph{hike} if
$p_t\ge(1+\theta)\tilde p_t$, with $\theta=5\%$. Event prices never enter $\mathcal W_t$, so a
promotion cannot drag down its own reference; the base attached to an event is the one
prevailing just before it began. An event lasting four periods is treated as a regime change
and re-seeds $\mathcal W_t$.

For each event $m$ with base $\tilde p_m$, pre-event units $\bar q^{\mathrm{pre}}_m$ (mean of
the last two regular periods) and post-start units $\bar q^{\mathrm{post}}_m$ (mean of the
first two event periods) at mean event price $p_m$:
\begin{equation}
\Delta^p_m=100\Big(\frac{p_m}{\tilde p_m}-1\Big),\qquad
L_m=100\Big(\frac{\bar q^{\mathrm{post}}_m}{\bar q^{\mathrm{pre}}_m}-1\Big),\qquad
\E_m=\max\!\Big(\frac{L_m}{\Delta^p_m},\,-10\Big).
\label{eq:event}
\end{equation}
Only \emph{directional} events (dip with lift, or hike with drop) receive an elasticity. If any
directional dip is significant ($|\Delta^p_m|\ge10\%$), all significant dips are used;
otherwise the eight deepest. The item's empirical elasticity is the depth-weighted mean
$\E^{\mathrm{emp}}_i=\sum_m|\Delta^p_m|\E_m/\sum_m|\Delta^p_m|$, clipped to $[-10,0]$.
It is then blended with the ML elasticity $\E^{\mathrm{ML}}_i$, read from the demand model's price
response with the elasticity equations and optimisers of~\cite{tgelastic2020} (clipped to $[-10,0]$):
\begin{equation}
\E_i=\begin{cases}\lambda\,\E^{\mathrm{emp}}_i+(1-\lambda)\,\E^{\mathrm{ML}}_i, & n^{\mathrm{sel}}_i\ge1,\ \E^{\mathrm{emp}}_i<0,\\
\E^{\mathrm{ML}}_i,& \text{otherwise,}\end{cases}\qquad \lambda=0.8 .
\label{eq:blend}
\end{equation}
Items left without an elasticity are filled from the analogue model (Sect.~\ref{sec:analogue}),
with hierarchy means as the legacy fallback; every item carries an \texttt{elasticity\_source} tag.

\textbf{Post-promo elasticity.} The same traversal yields the units in the regular periods
immediately \emph{after} each dip closes. Regressing their log-ratio to the pre-event level on
$\log(1+\Delta^p_m)$ gives the post-promo elasticity $\E^{\mathrm{post}}_i$, which captures
forward buying and pantry loading~\cite{vanheerde2000}.

\subsection{Promo-Time Causal Elasticity Curves}
\label{sec:promotime}
While a promotion runs, price cuts are confounded with trend and seasonality, so we fit a
depth-dependent curve with a two-stage partialling-out
estimator~\cite{frisch1933,lovell1963,chernozhukov2018}.

\textbf{Stage 1 (residualise).} Build a confounder matrix $W_{it}$ of trend, Fourier
seasonal terms and a promotion-intensity column, and residualise log units and log price
ratio: $\tilde y_{it}=\log q_{it}-\hat{\mathrm{E}}[\log q_{it}\mid W]$,
$\tilde x_{it}=\log(p_{it}/p^0_i)-\hat{\mathrm{E}}[\log(p_{it}/p^0_i)\mid W]$.

\textbf{Stage 2 (curve).} Fit, through the origin,
$\tilde y_{it}=b_{1i}\tilde x_{it}+b_{2i}\,d_{it}\tilde x_{it}+u_{it}$, which yields the
elasticity curve $e_i(d)=b_{1i}+b_{2i}d$ over discount depth $d$. The intercept $b_1$ is the
point elasticity at a shallow discount; a positive $b_2$ encodes \emph{saturation}: each
additional point of discount buys proportionally less volume. Weeks in which sales hit
available stock are right-censored; there the Gaussian likelihood is replaced by its survival
term, $\log\bar\Phi\big((\tilde y^{\mathrm{cap}}-\hat y)/\sigma\big)$, a Tobit
fit~\cite{tobin1958} that prevents stock-outs from understating elasticity. Without enough
distinct depths, $b_2$ is borrowed (Sect.~\ref{sec:analogue}).

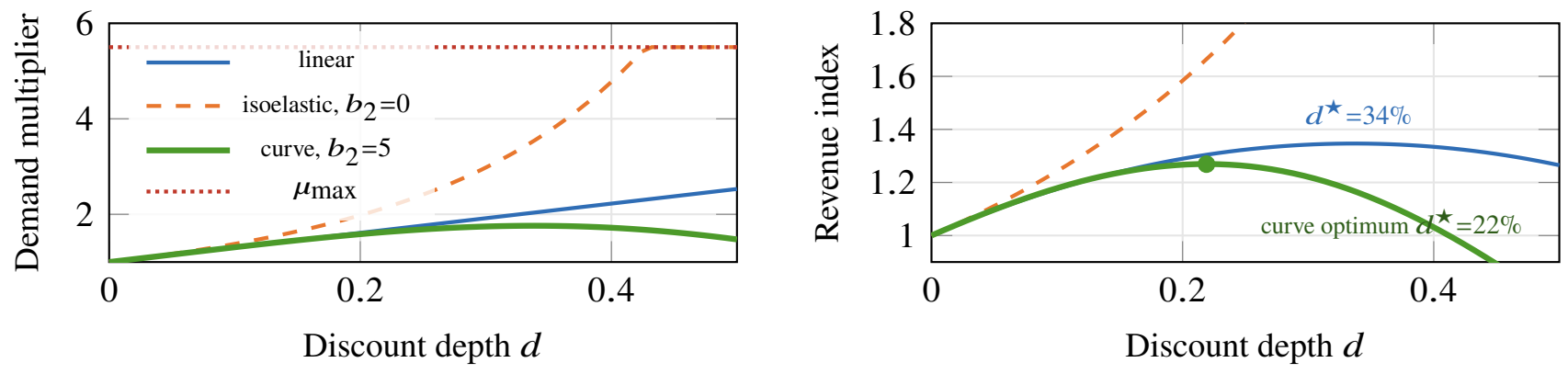
\begin{figure}[t]
\centering
\begin{tikzpicture}
\begin{groupplot}[group style={group size=2 by 1, horizontal sep=14mm},
  width=0.5\textwidth, height=3.3cm, xlabel={Discount depth $d$},
  tick label style={font=\scriptsize}, label style={font=\scriptsize},
  legend style={font=\tiny, draw=none, fill=white, fill opacity=0.8, text opacity=1},
  xmin=0, xmax=0.5, grid=major, grid style={gray!20}]
\nextgroupplot[ylabel={Demand multiplier}, ymin=1, ymax=6, legend pos=north west]
\addplot[tgblue, thick, domain=0:0.5, samples=60] {1+3.06*x}; \addlegendentry{linear}
\addplot[tgorange, thick, dashed, domain=0:0.5, samples=60] {min((1-x)^(-3.06),5.5)}; \addlegendentry{isoelastic, $b_2{=}0$}
\addplot[tggreen, very thick, domain=0:0.5, samples=60] {(1-x)^(-3.06+5*x)}; \addlegendentry{curve, $b_2{=}5$}
\addplot[tgred, densely dotted, thick, domain=0:0.5] {5.5}; \addlegendentry{$\mu_{\max}$}
\nextgroupplot[ylabel={Revenue index}, ymin=0.9, ymax=1.8, legend pos=north west]
\addplot[tgblue, thick, domain=0:0.5, samples=60] {(1+3.06*x)*(1-x)};
\addplot[tgorange, thick, dashed, domain=0:0.5, samples=60] {min((1-x)^(-2.06),5.5*(1-x))};
\addplot[tggreen, very thick, domain=0:0.5, samples=60] {(1-x)^(1-3.06+5*x)};
\addplot[only marks, mark=*, mark size=1.6pt, tggreen] coordinates {(0.219,1.269)};
\node[font=\tiny, anchor=south west, tggreen!60!black] at (axis cs:0.25,0.93) {curve optimum $d^\star{=}22\%$};
\node[font=\tiny, anchor=south, tgblue] at (axis cs:0.34,1.36) {$d^\star{=}34\%$};
\end{groupplot}
\end{tikzpicture}
\caption{Response models for $b_1=-3.06$ (mean of the 39-SKU brand of Sect.~\ref{sec:brand}). Left: demand multiplier.
Right: revenue index $(1-d)\,m(d)$. Constant isoelastic response is unbounded until the
cap binds; the curvature term $b_2$ produces a finite, interior optimum.}
\label{fig:curve}
\end{figure}

\textbf{Demand multiplier.} The curve is applied isoelastically on a grid of price changes,
\begin{equation}
m_i(c)=\mathrm{clip}\Big((1+c)^{\,e_i(d)},\ m_{\min},\ \mu_{\max}\Big),\qquad d=\max(-c,0),
\label{eq:mult}
\end{equation}
optionally snapped to a promotion price ladder (e.g.\ 10, 20, 30\% off).
Fig.~\ref{fig:curve} shows that only the curve has an interior revenue optimum.

\textbf{Bayesian promo-time update.} When the first \emph{clean} promotion week is realised
(depth $\ge5$ points, positive baseline, not inventory-bound, not a stock-out), the implied
elasticity is inverted from the realised lift,
$\hat e_i=\log(q^{\mathrm{act}}_{it}/\hat q_{it})/\log(1+c_{it})$, rejected if
$|\hat e_i-e^{\mathrm{prior}}_i|>4$, and combined as
$e^{\mathrm{upd}}=(1-\omega)e^{\mathrm{prior}}+\omega\hat e_i$ with $\omega=0.35$. The difference
$\delta_i=e^{\mathrm{upd}}-e^{\mathrm{prior}}$ is applied as a level shift to the whole curve,
preserving its curvature, and the remaining promotion weeks can be re-optimised.

\subsection{Learned Analogue Pooling for Point and Curve Elasticity}
\label{sec:analogue}
Most items have too few events for a precise own estimate, and this is where the chain begins.
Rather than averaging a fixed hierarchy group, we \emph{learn} which attributes make SKUs
analogous. Let $\hat\beta_i$ be the own estimate (either $b_1$ or $b_2$) with standard error $s_i$.

\textbf{Donors.} An item may lend only if $s_i\le s^{\max}$, it has enough observations and
real price variation, so a flat-priced item cannot spread a spurious estimate.

\textbf{Features.} For every item we build one-hot encodings of each hierarchy \emph{prefix}
(department, department$\,>\,$category, department$\,>\,$category$\,>\,$sub-category),
categorical attributes such as brand, retailer and colour, the standardised log typical price
\emph{and} its deviation from the item's group mean (the \emph{price line}), and a truncated-SVD
summary of TF-IDF vectors~\cite{salton1988} of free-text descriptions. Unusable columns are dropped automatically.

\textbf{Model.} A ridge regression~\cite{hoerl1970} maps features to donor elasticities with
precision weights $1/(s_i^2+\tau_0^2)$; with one block per hierarchy prefix, the penalty pulls
a thin sub-category towards its category and the category towards its department. The penalty $\alpha$ is chosen by $K$-fold cross-validation in which
folds keep all rows of one product together, and every donor's prediction is out-of-fold,
so no item's estimate enters its own prior. The same folds score the exact-group median, and
the reported \emph{skill} $1-\mathrm{MSE}_{\mathrm{model}}/\mathrm{MSE}_{\mathrm{group}}$
tells the planner whether borrowing adds anything.

\textbf{Shrinkage and serving.} Let $\mu_i$ be the prediction and $\tau^2$ the between-item
variance, i.e.\ the out-of-fold residual variance net of mean sampling variance, floored at a
trust share $\phi$ of $\mathrm{Var}(y)$. Then
\begin{equation}
\beta_i=\omega_i\hat\beta_i+(1-\omega_i)\mu_i,\qquad \omega_i=\frac{\tau^2}{\tau^2+s_i^2},\qquad
\mathrm{SE}_i=\big(s_i^{-2}+\tau^{-2}\big)^{-1/2},
\label{eq:shrink}
\end{equation}
with $\omega_i=0$ for items without an own estimate. An item is \emph{served} a borrowed value
only if its hierarchy is known, its own group contains at least $n^{\min}=3$ donors, and its
price lies within the donors' trimmed price range; otherwise no value is invented and the
abstention reason is logged. The identical procedure is run for $b_1$ and $b_2$: the borrowed
$b_1$ sets the \emph{level} of the item's curve and the borrowed $b_2$ sets how fast that level
\emph{flattens} with depth, so Eq.~\eqref{eq:shrink} feeds Eq.~\eqref{eq:mult} directly.

\subsection{Self-Price, Substitution and Post-Promo Effects}
\label{sec:cross}
Gross promotional lift overstates value because part of it is cannibalised from substitutes
and part is borrowed from future weeks~\cite{gupta1988,vanheerde2003}. For a promotion
$c_i<0$ held for $T$ weeks we write net incremental units as
\begin{equation}
\begin{split}
\Delta Q^{\mathrm{net}}_i(c)={}&\underbrace{T\hat q_i\big(m_i(c)-1\big)}_{\text{self-price lift}}
-\underbrace{T\!\!\sum_{j\in\mathcal N(i)}\!\!\hat q_j\big(1-(1+c)^{\E_{ji}}\big)}_{\text{substitution}}\\
&-\underbrace{\textstyle\sum_{k\ge1}\hat q_{i,T+k}\big(1-(1+c)^{-\E^{\mathrm{post}}_i\rho^{k-1}}\big)}_{\text{post-promo dip}} .
\end{split}
\label{eq:net}
\end{equation}
The neighbourhood $\mathcal N(i)$ is exactly the analogue group of Sect.~\ref{sec:analogue}:
items in the same sub-category and price line, which are the most plausible substitutes, so
the analogue model is reused a second time, now to bound cannibalisation. The cross elasticity
$\E_{ji}\ge0$ is estimated with the same Stage-1 residualisation, regressing $\tilde y_{jt}$ on
$\tilde x_{jt}$ (own) and $\tilde x_{it}$ (partner) jointly, and is pooled across pairs with pair
features (same brand, price ratio, description similarity). The decay $\rho\in(0,1)$ spreads
the dip over the weeks after the promotion ends. For the brand of Sect.~\ref{sec:brand}, $\E^{\mathrm{post}}\approx-0.80$ at its 26\% mean
discount puts the first post-promo week at $0.74^{0.80}=78\%$ of baseline, a 22\% dip.

\subsection{Optimisation, Overrides, Inventory and Trade Funding}
\label{sec:opt}
\textbf{Corridors.} Items are ranked by $\hat q_i$ into five demand buckets whose corridors are
scaled by $\{50,60,70,80,100\}\%$ from fastest to slowest, so best-sellers move least.
Candidates run from the floor in steps of $s$, with 0\% always included. The optimiser
evaluates Eq.~\eqref{eq:obj} with $Q_i(c)=\hat q_i+\Delta Q^{\mathrm{net}}_i(c)/T$ from
Eq.~\eqref{eq:net}, so the recommendation maximises \emph{net}, not gross, value.

\textbf{Overrides.} Business rules (column selectors plus a forced change) apply to the whole
series, last rule winning; forced series bypass the corridor and must be approved.

\textbf{Tactical inventory view.} The strategic change $c^{\mathrm{s}}_i$ is adjusted by stock risk:
\begin{equation}
c^{\mathrm{tac}}_i=\begin{cases}
-1.5\,|c^{\mathrm{s}}_i| & \text{excess stock (liquidate)},\\
\max(c^{\mathrm{s}}_i,0) & \text{stock-out risk with wastage, zero cover or lost sales},\\
c^{\mathrm{s}}_i & \text{otherwise.}\end{cases}
\label{eq:tac}
\end{equation}
\textbf{Trade funding.} With funding $f_i$ per unit (by default the full discount,
$f_i=p^0_i-p^{\mathrm s}_i$) and shipment price $p^{\mathrm{sh}}_i$, the incremental promotion profit
per day is $\Pi_i=\hat q_i m_i(c)\,(p^{\mathrm{sh}}_i-k_i-f_i)-\hat q_i\,(p^{\mathrm{sh}}_i-k_i)$;
promotions are ranked by $\Pi_i$ for the trade calendar. Algorithm~\ref{alg:tpo} summarises
the procedure.

\begin{algorithm}[!t]
\caption{Analogue-pooled, inventory-aware trade promotion optimisation}
\label{alg:tpo}
\footnotesize
\begin{algorithmic}[1]
\Require history $\{q_{it},p_{it}\}$, forecast $\hat q_{it}$, ML elasticity $\E^{\mathrm{ML}}$, stock, funding, settings
\Ensure strategic and tactical change, price, units, revenue, margin, $\Pi_i$ per item
\For{each item $i$ \textbf{in parallel}}
  \State flag promotions; $p^0_i\gets$ client regular price or ModeAboveMedian(365 days)
  \State events $\gets$ \Call{Traverse}{$q_i,p_i$}; select significant directional dips (else top-8)
  \State $\E^{\mathrm{emp}}_i\gets$ depth-weighted mean; $\E_i\gets$ Eq.~\eqref{eq:blend}; $\E^{\mathrm{post}}_i\gets$ post-event regression
  \State promo time: $(\hat b_{1i},\hat b_{2i},s_{1i},s_{2i})\gets$ FWL residualise + Tobit stage 2
\EndFor
\For{target $\beta\in\{b_1,b_2\}$}  \Comment{analogue pooling, Sect.~\ref{sec:analogue}}
  \State donors $\gets\{i: s_i\le s^{\max},\ \text{price variation ok}\}$; build hierarchy/brand/price-line/text features
  \State choose $\alpha$ by grouped CV; record skill vs.\ group median; fit ridge; $\mu_i\gets$ out-of-fold prediction
  \State \textbf{if} served$(i)$ \textbf{then} $\beta_i\gets$ Eq.~\eqref{eq:shrink} \textbf{else} keep own / hierarchy fill, log reason
\EndFor
\State apply Bayesian level shift from the first clean realised promotion week
\For{each item $i$ \textbf{in parallel}}
  \State bucket corridor $\mathcal C_i$; \textbf{for} $c\in\mathcal C_i$: $m_i(c)\gets$ Eq.~\eqref{eq:mult}; $Q_i(c)\gets$ Eq.~\eqref{eq:net}; $J_i(c)\gets$ Eq.~\eqref{eq:obj}
  \State $c^{\mathrm s}_i\gets\arg\max_c J_i(c)$ (first on ties); apply overrides
  \State $c^{\mathrm{tac}}_i\gets$ Eq.~\eqref{eq:tac}; round price points; compute $\Pi_i$
\EndFor
\end{algorithmic}
\end{algorithm}

\subsection{The Chain on One SKU: From Analogues to Net Lift}
\label{sec:chain}
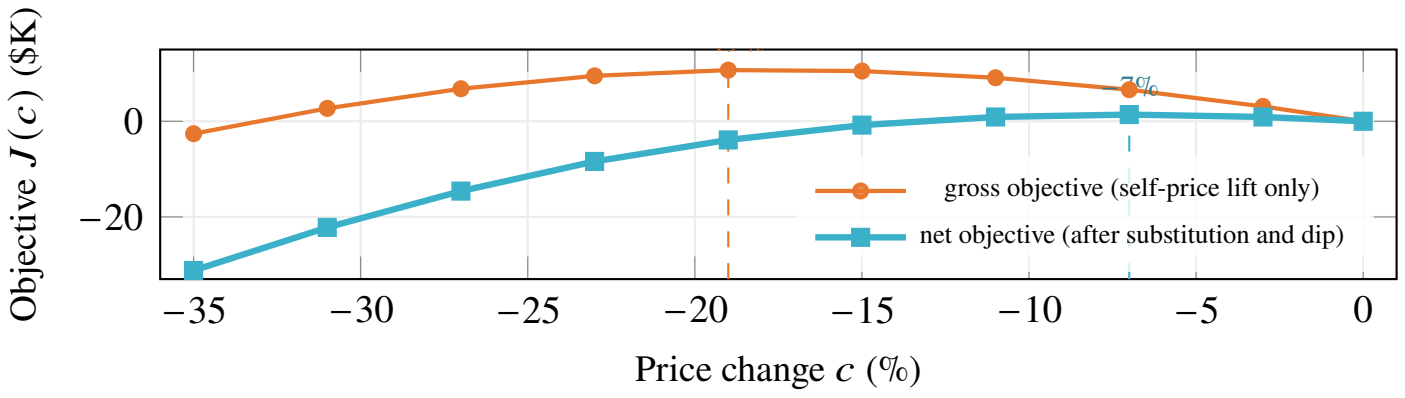
\begin{figure}[!t]
\centering
\begin{tikzpicture}
\begin{axis}[width=0.8\textwidth, height=3.1cm, xlabel={Price change $c$ (\%)}, ylabel={Objective $J(c)$ (\$K)},
  tick label style={font=\scriptsize}, label style={font=\scriptsize}, xmin=-36, xmax=1, ymin=-33, ymax=15,
  legend style={font=\tiny, at={(0.98,0.03)}, anchor=south east, draw=none, fill=white, fill opacity=0.8, text opacity=1},
  grid=major, grid style={gray!15}]
\addplot[tgorange, thick, mark=*, mark size=1.2pt] coordinates {(-35,-2.6)(-31,2.7)(-27,6.8)(-23,9.5)(-19,10.7)(-15,10.5)(-11,9.1)(-7,6.6)(-3,3.1)(0,0)};
\addlegendentry{gross objective (self-price lift only)}
\addplot[tgteal, very thick, mark=square*, mark size=1.2pt] coordinates {(-35,-31.2)(-31,-22.2)(-27,-14.6)(-23,-8.4)(-19,-3.9)(-15,-0.8)(-11,0.9)(-7,1.4)(-3,0.9)(0,0)};
\addlegendentry{net objective (after substitution and dip)}
\draw[tgorange, dashed] (axis cs:-19,-33) -- (axis cs:-19,10.7);
\draw[tgteal, dashed] (axis cs:-7,-33) -- (axis cs:-7,1.4);
\node[font=\tiny, tgorange!80!black, anchor=south] at (axis cs:-19,11.0) {$-19\%$};
\node[font=\tiny, tgteal!70!black, anchor=south] at (axis cs:-7,1.7) {$-7\%$};
\end{axis}
\end{tikzpicture}
\caption{Objective of Eq.~\eqref{eq:obj} over the candidate grid for the SKU of
Table~\ref{tab:chain}. Netting out substitution and the post-promo dip moves the recommended
discount from $-19\%$ to $-7\%$.}
\label{fig:netobj}
\end{figure}

Table~\ref{tab:chain} traces one illustrative SKU through every link of Fig.~\ref{fig:chain},
using the medians of the brand in Sect.~\ref{sec:brand}: \$27.00 regular price, \$8.52 unit
cost, 700 units a week and no usable price event of its own. Its sub-category holds five
donors, so it is served $b_1=-2.90$ (the brand median) and $b_2=5.0$; the curve flattens to
$-2.55$ at 7\% off and $-1.95$ at 19\% off. Over a four-week promotion with $w=50$, a
gross-only objective picks $-19\%$; the net objective ($\E_{ji}=0.25$ on 2,100 weekly
substitute units, $\E^{\mathrm{post}}=-0.80$, $\rho=0.5$) picks only $-7\%$
(Fig.~\ref{fig:netobj}). At $-7\%$ the item gains +20.3\% units gross and +12.3\% net at
break-even net margin; the gross optimum would have lost \$7.2K (13.9\%) of net margin, the
gross-positive, net-negative pattern the brand shows in Table~\ref{tab:brand}.

\begin{table}[!t]
\centering
\caption{The chain on one illustrative SKU (four-week promotion; baseline 2,800 units,
\$75.6K revenue, \$51.7K margin).}\label{tab:chain}
{\tablesetup
\begin{tabular*}{\textwidth}{@{\extracolsep{\fill}}clr@{}}
\toprule
Step & Link of the chain & Value \\
\midrule
1 & Analogue SKUs: donors in sub-category, own weight $\omega_i$ & 5 donors, $\omega_i=0$ \\
2 & Point elasticity $b_1$ (borrowed) & $-2.90$ \\
3 & Curve $e(d)=b_1+b_2d$ with borrowed $b_2=5.0$ & $e(0.07)=-2.55$ \\
4 & Recommendation: gross objective / net objective & $-19\%$ / $-7\%$ \\
5 & Promotional lift at $-7\%$: units / revenue / margin & +20.3\% / +11.9\% / +8.0\% \\
6 & Net lift after substitution and post-promo dip & +12.3\% / +3.8\% / 0.0\% \\
\bottomrule
\end{tabular*}}
\end{table}

\section{Experiments}
\label{sec:exp}
\subsection{Studies and Setup}
\emph{Study A} is a four-week promotion experiment on one consumer brand (anonymised as
Brand~A): all 39 of its SKUs were optimised jointly on the planning platform, with weekly
history from October 2024 to October 2026 and the promotion window starting in October 2026.
The corridor was configured for discount-only promotions ($c^{\max}{=}0$,
$c^{\min}{=}{-}35\%$, step 4), so the bucket-scaled floors are $-35$, $-28$, $-24$, $-21$ and
$-17\%$, and funding defaults to the full discount. \emph{Study B} is a controlled simulation
of the first link of the chain, analogue pooling (Sect.~\ref{sec:sim}). Table~\ref{tab:data}
summarises the Study A data.

\begin{table}[t]
\centering
\caption{Study A data: 39 Brand~A SKUs, four-week promotion window.}\label{tab:data}
{\tablesetup
\begin{tabular*}{\textwidth}{@{\extracolsep{\fill}}lrrrrr@{}}
\toprule
Quantity & Mean & Std & 25\% & Median & 75\% \\
\midrule
Regular price (\$) & 27.50 & 9.64 & 19.98 & 27.00 & 35.00 \\
Unit COGS (\$) & 8.76 & 3.10 & 6.23 & 8.52 & 10.73 \\
Self-price elasticity $\E_i$ & $-3.06$ & 1.95 & $-4.76$ & $-2.90$ & $-1.53$ \\
Recommended change (\%) & $-21.5$ & 11.9 & $-31$ & $-24$ & $-17$ \\
Promoted price (\$) & 21.47 & 7.90 & 15.61 & 19.99 & 24.95 \\
Unit lift (\%) & 84.1 & 71.2 & 26.1 & 73.3 & 132.1 \\
Revenue lift (\%) & 37.6 & 37.7 & 4.6 & 28.8 & 66.6 \\
Funding per unit (\$) & 6.03 & 4.18 & 3.17 & 5.60 & 9.45 \\
\bottomrule
\end{tabular*}}
\end{table}

\subsection{Study A: Four-Week Brand Promotion Experiment}
\label{sec:brand}
\begin{figure}[t]
\centering
\begin{tikzpicture}
\begin{groupplot}[group style={group size=1 by 2, vertical sep=3mm, x descriptions at=edge bottom},
  width=\textwidth, xmin=0, xmax=128,
  xtick={0,17,35,56,73,95,113}, xticklabels={Oct 24,Feb 25,Jun 25,Nov 25,Mar 26,Aug 26,Dec 26},
  tick label style={font=\scriptsize}, label style={font=\scriptsize}, grid=major, grid style={gray!15}]
\nextgroupplot[height=3.1cm, ylabel={Units (K)}, ymin=7, ymax=22, ytick={10,15,20},
  legend style={font=\tiny, at={(0.5,1.03)}, anchor=south, legend columns=4, draw=none}]
\addplot[tgblue, thick, mark=*, mark size=0.6pt] coordinates {(0,16.5)(1,18.5)(2,16)(3,14.5)(4,13.5)(5,15.5)(6,14.8)(7,15.3)(8,13.5)(9,15.2)(10,16)(11,12)(12,14.8)(13,15.5)(14,16.2)(15,13.5)(16,13.5)(17,16)(18,15)(19,17.2)(20,20)(21,19)(22,18.5)(23,17.5)(24,16.5)(25,16.5)(26,17)(27,16.8)(28,16)(29,16.5)(30,15.5)(31,14.8)(32,14.5)(33,14.5)(34,15.3)(35,16.2)(36,14.5)(37,13.5)(38,17.2)(39,17)(40,15.5)(41,15)(42,15)(43,14.8)(44,14)(45,14)(46,13.8)(47,15.5)(48,16)(49,17.5)(50,18.5)(51,16.2)(52,14.8)(53,16.2)(54,14.5)(55,15)(56,14.5)(57,14.5)(58,14)(59,13)(60,12.5)(61,13.2)(62,17)(63,20.2)(64,19.3)(65,15.5)(66,14.8)(67,18.8)(68,16.4)(69,16.5)(70,17.8)(71,17.6)(72,18.4)(73,19.5)(74,17)(75,18.5)(76,16.5)(77,16.3)(78,16)(79,15.2)(80,14.5)(81,14.3)(82,13.5)(83,14.5)(84,15.5)(85,15.5)(86,15.2)(87,15.6)(88,14.2)(89,14.5)(90,14.5)(91,14.2)(92,13.2)(93,14)(94,12.6)(95,12.2)(96,11.5)(97,10.4)(98,9.3)(99,9.4)(100,11.2)(101,12.0)(102,12.5)}; \addlegendentry{history}
\addplot[tggreen, thick, mark=*, mark size=0.6pt] coordinates {(102,12.5)(103,12.2)(104,11.6)(105,11.0)(106,10.6)(107,11.4)(108,11.3)(109,11.2)(110,10.8)(111,11.5)(112,10.9)(113,10.6)(114,10.7)(115,12.0)(116,13.8)(117,14.5)(118,13.6)(119,13.6)(120,13.8)(121,14.6)(122,14.8)(123,14.8)(124,14.5)(125,13.8)(126,12.0)}; \addlegendentry{baseline forecast}
\addplot[tgorange, thick, dashed, mark=*, mark size=1pt] coordinates {(103,20.2)(104,19.0)(105,17.6)(106,18.6)}; \addlegendentry{4-week promotion}
\addplot[tgteal, thick, densely dotted, mark=triangle*, mark size=1.2pt] coordinates {(106,18.6)(107,8.9)(108,9.6)(109,10.2)(110,11.3)}; \addlegendentry{post-promo dip}
\draw[gray, dashed] (axis cs:102.5,7) -- (axis cs:102.5,22);
\nextgroupplot[height=2.5cm, ylabel={Price (\$)}, ymin=15, ymax=34, ytick={20,30}]
\addplot[tggold, thick, mark=*, mark size=0.5pt] coordinates {(0,30.0)(1,30.3)(2,30.6)(3,30.1)(4,30.4)(5,30.0)(6,30.3)(7,30.6)(8,30.1)(9,30.4)(10,30.0)(11,30.3)(12,30.6)(13,30.1)(14,30.4)(15,30.0)(16,30.3)(17,30.6)(18,30.1)(19,30.4)(20,30.0)(21,30.3)(22,30.6)(23,30.1)(24,30.4)(25,30.0)(26,30.3)(27,30.6)(28,30.1)(29,30.4)(30,30.0)(31,30.3)(32,30.6)(33,26.0)(34,26.5)(35,27.0)(36,27.5)(37,28.0)(38,28.5)(39,30.4)(40,30.0)(41,30.3)(42,30.6)(43,30.1)(44,30.4)(45,30.0)(46,30.3)(47,30.6)(48,30.1)(49,30.4)(50,30.0)(51,30.3)(52,30.6)(53,30.1)(54,30.4)(55,30.0)(56,30.3)(57,30.6)(58,30.1)(59,30.4)(60,30.0)(61,30.3)(62,30.6)(63,30.1)(64,30.4)(65,30.0)(66,30.3)(67,30.6)(68,30.1)(69,30.4)(70,27.0)(71,27.0)(72,27.0)(73,27.0)(74,30.4)(75,30.0)(76,30.3)(77,30.6)(78,30.1)(79,30.4)(80,30.0)(81,30.3)(82,30.6)(83,30.1)(84,30.4)(85,30.0)(86,30.3)(87,30.6)(88,30.1)(89,30.4)(90,30.0)(91,30.3)(92,30.6)(93,30.1)(94,30.4)(95,30.0)(96,26.5)(97,26.5)(98,26.5)(99,26.5)(100,30.0)(101,25.0)(102,25.0)};
\addplot[tggold, thick, mark=*, mark size=0.5pt] coordinates {(102,29.6)(103,29.6)(104,29.6)(105,29.6)(106,29.6)(107,29.6)(108,29.6)(109,29.6)(110,29.6)(111,29.6)(112,29.6)(113,29.6)(114,29.6)(115,29.6)(116,29.6)(117,29.6)(118,29.6)(119,29.6)(120,29.6)(121,29.6)(122,29.6)(123,29.6)(124,29.6)(125,29.6)(126,29.6)};
\addplot[tgorange, thick, mark=*, mark size=1pt] coordinates {(103,22.0)(104,22.0)(105,22.0)(106,22.0)};
\draw[gray, dashed] (axis cs:102.5,15) -- (axis cs:102.5,34);
\end{groupplot}
\end{tikzpicture}
\caption{Study A, Brand~A weekly units (top) and average price (bottom), redrawn from the
platform scenario view. The dashed line marks the start of the four-week promotion.}
\label{fig:brand}
\end{figure}
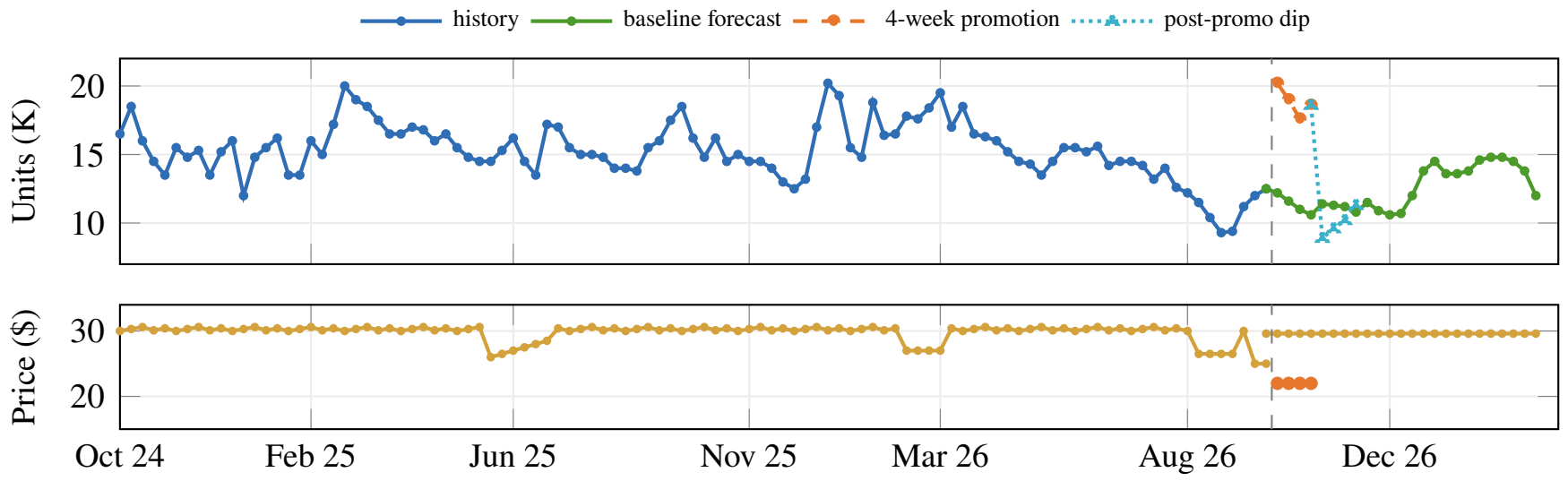

\textbf{Elasticity and recommendations.} The brand is price-responsive on average (mean
$\E_i=-3.06$, median $-2.90$, 33 of 39 SKUs with $\E_i<-1$). The system discounted 32 SKUs (mean
depth 26.2\%) and left 7 at full price, all with $\E_i\ge-1.05$, where a cut barely raises
revenue. The 12 SKUs with $\E_i\le-4$ (mean $-5.48$) get the deepest cuts (mean $-31.6\%$),
and 12 SKUs sit on the $-31\%$ or $-35\%$ floors, because the plan's linear response $Q=\hat q(1+\E c)$, which
the lifts in Table~\ref{tab:skus} follow exactly, gives them no reason to stop short of the
floor, the corner-solution pattern that the curvature $b_2$ of Fig.~\ref{fig:curve} removes.
Fig.~\ref{fig:brand} shows the brand's weekly trajectory: the average price drops from about
\$29.6 to \$22 for four weeks, units rise well above the baseline forecast, and after the
promotion units fall about 21\% below baseline before recovering within roughly four weeks.

\textbf{Brand-level outcome.} Table~\ref{tab:brand} reports the outcome. The platform reports +10.9K units (+20.64\%), +\$76.7K revenue (+6.87\%) and
+\$2.9K margin (+0.37\%) over the four weeks at a mean elasticity of $-3.06$. Weighting each SKU by baseline volume (non-promotion profit over unit margin) reproduces
the platform totals closely (+23.6\% units, +7.0\% revenue, margin $-1.2\%$) and gives a
volume-weighted elasticity of only $-1.46$.

\begin{table}[t]
\centering
\caption{Study A, four-week Brand~A outcome: promotional (gross) lift reported by the platform,
its volume-weighted SKU reconstruction, and net lift after the post-promo dip estimated from
Fig.~\ref{fig:brand} (16\% of the bump, $\E^{\mathrm{post}}\approx-0.80$).}\label{tab:brand}
{\tablesetup
\begin{tabular*}{\textwidth}{@{\extracolsep{\fill}}lrrrr@{}}
\toprule
Metric & Platform (gross) & Vol.-weighted SKUs & Post-promo dip & Net lift \\
\midrule
Units & +10.9K (+20.64\%) & +23.6\% & $-1.7$K & +9.2K (+17.3\%) \\
Revenue & +\$76.7K (+6.87\%) & +7.0\% & $-\$36.9$K & +\$39.8K (+3.6\%) \\
Margin & +\$2.9K (+0.37\%) & $-1.2\%$ & $-\$25.9$K & $-\$23.0$K ($-2.9\%$) \\
Elasticity & $-3.06$ (mean) & $-1.46$ & $\approx-0.80$ & \\
\bottomrule
\end{tabular*}}
\end{table}

\textbf{Where the volume is.} Fig.~\ref{fig:skus} orders the 39 SKUs by baseline volume. Five
SKUs carry 83.1\% of the brand's volume, and all five are inelastic (volume-weighted
$\E=-1.16$). Two are left at full price and three receive shallow, \emph{interior} cuts
($-9\%$, $-13\%$, $-21\%$), since for an inelastic item a deeper cut destroys margin faster than
it adds units. The remaining 34 SKUs (16.9\% of volume, weighted $\E=-2.90$) are cut by about
20\% and lift units by 71.8\% and revenue by 27.1\%. Table~\ref{tab:skus} shows the five large SKUs add only
13.7\% units and 2.6\% revenue at a 2.5\% margin loss, so the brand total stays modest.

\begin{figure}[t]
\centering
\begin{tikzpicture}
\begin{axis}[width=0.86\textwidth, height=3.1cm, ybar, bar width=3.2pt, ymin=0, ymax=260,
  xmin=-1, xmax=39, xlabel={SKUs ordered by baseline volume (largest first)}, ylabel={Unit lift (\%)},
  xtick={0,4,9,14,19,24,29,34,38}, xticklabels={1,5,10,15,20,25,30,35,39},
  tick label style={font=\scriptsize}, label style={font=\scriptsize}, axis y line*=left]
\addplot[fill=tgorange!70, draw=tgorange!80!black] coordinates {(0,18.4) (1,10.7) (2,0.0) (3,41.7) (4,0.0) (5,132.4) (6,0.0) (7,46.3) (8,75.6) (9,32.0) (10,0.0) (11,73.3) (12,76.7) (13,69.5) (14,118.8) (15,55.6) (16,87.8) (17,131.9) (18,52.4) (19,186.3) (20,0.0) (21,0.0) (22,96.0) (23,197.4) (24,242.1) (25,121.4) (26,22.6) (27,149.4) (28,0.0) (29,150.8) (30,233.3) (31,178.9) (32,117.2) (33,31.5) (34,76.1) (35,175.4) (36,184.7) (37,29.6) (38,64.3)};
\end{axis}
\begin{axis}[width=0.86\textwidth, height=3.1cm, ymin=0, ymax=105, xmin=-1, xmax=39,
  axis y line*=right, axis x line=none, ylabel={Cumulative volume (\%)},
  tick label style={font=\scriptsize}, label style={font=\scriptsize}]
\addplot[tgblue, thick, mark=*, mark size=0.8pt] coordinates {(0,25.0) (1,47.7) (2,62.3) (3,72.8) (4,83.1) (5,86.5) (6,88.9) (7,90.2) (8,91.4) (9,92.4) (10,93.4) (11,94.3) (12,95.2) (13,96.0) (14,96.7) (15,97.5) (16,98.2) (17,98.7) (18,99.1) (19,99.4) (20,99.5) (21,99.6) (22,99.6) (23,99.7) (24,99.7) (25,99.8) (26,99.8) (27,99.8) (28,99.9) (29,99.9) (30,99.9) (31,99.9) (32,99.9) (33,99.9) (34,100.0) (35,100.0) (36,100.0) (37,100.0) (38,100.0)};
\draw[tgblue, dashed] (axis cs:4,0) -- (axis cs:4,83.1);
\node[font=\tiny, tgblue, anchor=west, fill=white, inner sep=1pt] at (axis cs:5,55) {top 5 SKUs = 83.1\% of volume};
\end{axis}
\end{tikzpicture}
\caption{Study A. Bars: recommended unit lift per SKU; line: cumulative share of baseline
volume. Volume sits in SKUs with small lifts; large lifts sit in SKUs with little volume.}
\label{fig:skus}
\end{figure}
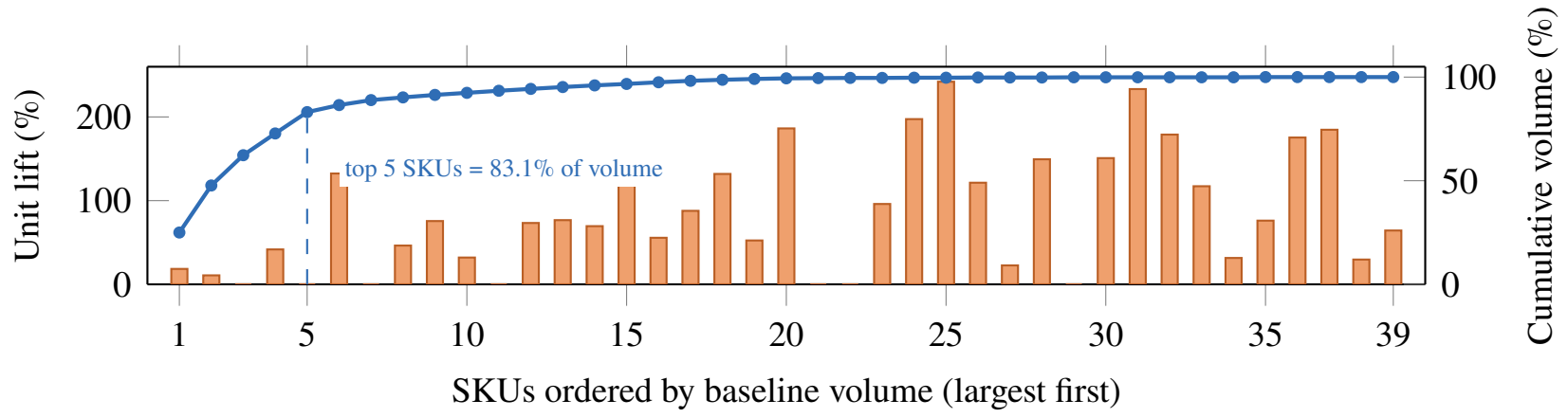
\begin{table}[t]
\centering
\caption{Study A, SKU-level evidence: the five largest SKUs and the remaining 34 (volume
weighted). $\Pi$ is incremental promotion profit per day under full funding.}\label{tab:skus}
{\tablesetup
\begin{tabular*}{\textwidth}{@{\extracolsep{\fill}}lrrrrrrr@{}}
\toprule
SKU & Volume & $\E$ & Change & Units & Revenue & Margin & $\Pi$ (\$/day) \\
\midrule
\#2 & 25.0\% & $-1.42$ & $-13\%$ & 18.4\% & 3.0\% & $-4.2\%$ & $-731$ \\
\#1 & 22.7\% & $-1.19$ & $-9\%$ & 10.7\% & 0.8\% & $-3.8\%$ & $-720$ \\
\#3 & 14.6\% & $-0.33$ & 0\% & 0.0\% & 0.0\% & 0.0\% & 0 \\
\#4 & 10.5\% & $-1.99$ & $-21\%$ & 41.7\% & 12.0\% & $-1.6\%$ & $-574$ \\
\#5 & 10.3\% & $-0.82$ & 0\% & 0.0\% & 0.0\% & 0.0\% & 0 \\
\midrule
Top 5 & 83.1\% & $-1.16$ & $-9.3\%$ & 13.7\% & 2.6\% & $-2.5\%$ & $-2{,}025$ \\
Other 34 & 16.9\% & $-2.90$ & $-20.3\%$ & 71.8\% & 27.1\% & 4.8\% & $-1{,}593$ \\
\bottomrule
\end{tabular*}}
\end{table}

\textbf{From promotional lift to net lift.} The platform reports gross lift over the promotion
window. We net out the post-promo dip from the trajectory in Fig.~\ref{fig:brand}: units in the
first post-promo week are 21\% below baseline, which at the brand's $\$22/\$29.6$ price ratio
implies $\E^{\mathrm{post}}\approx-0.80$, and over four post-promo weeks the dip equals 16\% of
the promotional bump. Applying that share to the platform's gross lift, at the brand's average
revenue (\$21.14) and margin (\$14.84) per baseline unit, gives the net column of Table~\ref{tab:brand}: net units
fall to +17.3\%, net revenue to +3.6\%, and the margin gain of \$2.9K turns into a loss of about
\$23.0K. Within-brand substitution is not netted here, so these net figures are an upper bound.

\textbf{Funding and guardrails.} Under full funding the promotion costs the brand \$3.6K of
profit per day against \$10.1K of non-promotion profit, a 35.6\% reduction, and only 2 of 39
SKUs have positive incremental profit. One of them, SKU~\#25, shows why the planner checks
matter: at $-31\%$ its promotional price (\$4.14) falls below its unit cost (\$5.82), giving a
$-2{,}204\%$ margin lift that the cost guardrail flags. As in Sect.~\ref{sec:chain}, a gross-positive
promotion turns net-negative once the dip and funding are counted.

\subsection{Study B: Does Analogue Pooling Recover Elasticity?}
\label{sec:sim}
True elasticities are unobservable on real data, so we test the first link in simulation:
1,440 SKUs in a $4\times3\times4$ hierarchy of departments, categories and sub-categories with
12 brands. True $b_1$ combines hierarchy and brand effects, a price-line slope and noise; true
$b_2$ varies by sub-category. With $k$ Poisson price events per SKU, own estimates have standard error $1.1/\sqrt{k}$ for $b_1$ (23.6\% of SKUs have none) and
$3/\sqrt{k}$ for $b_2$ (identified only when $k\ge3$; 61\% have none). Over 20 seeds we compare
own-only (zero when absent), hierarchy-mean fill, group donor median and the learned pool,
reporting RMSE and revenue \emph{regret}: the revenue gain lost (points of baseline) by choosing
$c\in[-35,0]$ with the estimated instead of the true elasticity.

\begin{table}[!htb]
\centering
\caption{Study B, mean $\pm$ s.d.\ over 20 seeds. RMSE on all SKUs, on borrowers (no own
estimate) and on noisy own estimates ($s>0.6$); regret in revenue-gain points.}\label{tab:sim}
{\tablesetup
\begin{tabular*}{\textwidth}{@{\extracolsep{\fill}}lcccc@{}}
\toprule
Metric & Own only & Hierarchy fill & Group median & \textbf{Learned pool} \\
\midrule
$b_1$ RMSE, all & 1.589 $\pm$ 0.195 & 0.710 $\pm$ 0.024 & 0.718 $\pm$ 0.025 & \textbf{0.366} $\pm$ 0.008 \\
$b_1$ RMSE, borrowers & 2.969 $\pm$ 0.397 & 0.545 $\pm$ 0.046 & 0.587 $\pm$ 0.053 & \textbf{0.416} $\pm$ 0.022 \\
$b_1$ RMSE, noisy own & 0.884 $\pm$ 0.040 & 0.884 $\pm$ 0.040 & 0.884 $\pm$ 0.040 & \textbf{0.372} $\pm$ 0.016 \\
Regret (points) & 8.18 $\pm$ 1.89 & 1.14 $\pm$ 0.25 & 1.17 $\pm$ 0.26 & \textbf{0.31} $\pm$ 0.13 \\
$b_2$ RMSE, all & 1.932 $\pm$ 0.095 & 0.978 $\pm$ 0.030 & 1.007 $\pm$ 0.037 & \textbf{0.598} $\pm$ 0.033 \\
\bottomrule
\end{tabular*}}
\end{table}

Table~\ref{tab:sim} shows that the learned pool halves the overall $b_1$ RMSE (0.71 to 0.37)
and cuts revenue regret by 73\% against the hierarchy fill: brand and price line cut borrower
error by 24\%, and shrinkage (Eq.~\eqref{eq:shrink}) cuts noisy-estimate error by 58\%. For $b_2$ the gain comes from shrinking noisy own curvatures (0.98 to 0.60); on pure
borrowers the model only matches the hierarchy fill (skill $+0.07$).

\section{Conclusion}
We presented a TPO system organised as a single chain. Analogue SKUs, selected by a learned
model over category, brand, sub-category, price line and description, supply the point
elasticity and the curvature where an item's own evidence is weak; these form an elasticity
curve measured at promo time and post-promo time; the curve gives the optimiser an interior
price recommendation; and the recommendation's promotional lift is netted for substitution and
post-promo dips into the lift the business keeps. In a four-week experiment on a 39-SKU brand
(mean elasticity $-3.06$) the plan lifts units by 20.6\% and revenue by 6.9\% at flat margin,
the volume-weighted elasticity is only $-1.46$, and netting the post-promo dip turns the margin
gain negative; in simulation, analogue pooling halves elasticity error. Future work will
measure realised against planned lift in live promotions.

\begin{credits}
\subsubsection{\ackname} The authors would like to thank Ankit Raj and Vishnu Bhanderi for their contribution towards Elasticity architecture improvements; Ankit Solanki and Vasu Bhanderi for front development; Siddharth Shahi and Sharad LM for helpful discussions and insights over the methodology and implementation.
\end{credits}
%
%


\begin{thebibliography}{99}
\bibitem{ailawadi1998}
Ailawadi, K.L., Neslin, S.A.: The effect of promotion on consumption: Buying more and consuming it faster. Journal of Marketing Research \textbf{35}(3), 390-398 (1998)

\bibitem{baardman2018}
Baardman, L., Levin, I., Perakis, G., Singhvi, D.: Leveraging comparables for new product sales forecasting. Production and Operations Management \textbf{27}(12), 2340-2343 (2018)

\bibitem{bijmolt2005}
Bijmolt, T.H.A., van Heerde, H.J., Pieters, R.G.M.: New empirical generalizations on the determinants of price elasticity. Journal of Marketing Research \textbf{42}(2), 141-156 (2005)

\bibitem{blattberg1990}
Blattberg, R.C., Neslin, S.A.: Sales Promotion: Concepts, Methods, and Strategies. Prentice Hall, Englewood Cliffs (1990)

\bibitem{chernozhukov2018}
Chernozhukov, V., Chetverikov, D., Demirer, M., Duflo, E., Hansen, C., Newey, W., Robins, J.: Double/debiased machine learning for treatment and structural parameters. The Econometrics Journal \textbf{21}(1), C1-C68 (2018)

\bibitem{cohen2017}
Cohen, M.C., Leung, N.H.Z., Panchamgam, K., Perakis, G., Smith, A.: The impact of linear optimization on promotion planning. Operations Research \textbf{65}(2), 446-468 (2017)

\bibitem{efron1973}
Efron, B., Morris, C.: Stein's estimation rule and its competitors - an empirical Bayes approach. Journal of the American Statistical Association \textbf{68}(341), 117-130 (1973)

\bibitem{ferreira2016}
Ferreira, K.J., Lee, B.H.A., Simchi-Levi, D.: Analytics for an online retailer: Demand forecasting and price optimization. Manufacturing \& Service Operations Management \textbf{18}(1), 69-88 (2016)

\bibitem{frisch1933}
Frisch, R., Waugh, F.V.: Partial time regressions as compared with individual trends. Econometrica \textbf{1}(4), 387-401 (1933)

\bibitem{gelman2007}
Gelman, A., Hill, J.: Data Analysis Using Regression and Multilevel/Hierarchical Models. Cambridge University Press, Cambridge (2007)

\bibitem{gupta1988}
Gupta, S.: Impact of sales promotions on when, what, and how much to buy. Journal of Marketing Research \textbf{25}(4), 342-355 (1988)

\bibitem{hoerl1970}
Hoerl, A.E., Kennard, R.W.: Ridge regression: Biased estimation for nonorthogonal problems. Technometrics \textbf{12}(1), 55-67 (1970)

\bibitem{kok2007}
K\"ok, A.G., Fisher, M.L.: Demand estimation and assortment optimization under substitution: Methodology and application. Operations Research \textbf{55}(6), 1001-1021 (2007)

\bibitem{lovell1963}
Lovell, M.C.: Seasonal adjustment of economic time series and multiple regression analysis. Journal of the American Statistical Association \textbf{58}(304), 993-1010 (1963)

\bibitem{montgomery1997}
Montgomery, A.L.: Creating micro-marketing pricing strategies using supermarket scanner data. Marketing Science \textbf{16}(4), 315-337 (1997)

\bibitem{salton1988}
Salton, G., Buckley, C.: Term-weighting approaches in automatic text retrieval. Information Processing \& Management \textbf{24}(5), 513-523 (1988)

\bibitem{smith1998}
Smith, S.A., Achabal, D.D.: Clearance pricing and inventory policies for retail chains. Management Science \textbf{44}(3), 285-300 (1998)

\bibitem{talluri2004}
Talluri, K.T., van Ryzin, G.J.: The Theory and Practice of Revenue Management. Springer, New York (2004)

\bibitem{tgdemand2020}
Verma,Ankur: Consumer behaviour in retail: next logical purchase using deep neural network. arXiv preprint arXiv:2010.06952 (2020), \url{https://arxiv.org/abs/2010.06952}

\bibitem{tgelastic2020}
Verma,Ankur: Offer Personalization using Temporal Convolution Network and Optimization. arXiv preprint arXiv:2010.08130 (2020), \url{https://arxiv.org/abs/2010.08130}

\bibitem{tobin1958}
Tobin, J.: Estimation of relationships for limited dependent variables. Econometrica \textbf{26}(1), 24-36 (1958)

\bibitem{vanheerde2000}
van Heerde, H.J., Leeflang, P.S.H., Wittink, D.R.: The estimation of pre- and postpromotion dips with store-level scanner data. Journal of Marketing Research \textbf{37}(3), 383-395 (2000)

\bibitem{vanheerde2003}
van Heerde, H.J., Gupta, S., Wittink, D.R.: Is 75\% of the sales promotion bump due to brand switching? No, only 33\% is. Journal of Marketing Research \textbf{40}(4), 481-491 (2003)
\end{thebibliography}
\end{document}